\documentclass{article}
\usepackage{arxiv}
\usepackage[utf8]{inputenc} % allow utf-8 input
\usepackage[T1]{fontenc}    % use 8-bit T1 fonts
\usepackage{hyperref, enumitem, url, amsmath, amssymb, amsthm, booktabs, float, amsfonts, subcaption, microtype, graphicx, natbib, doi, algorithm, algpseudocode}
\numberwithin{equation}{section}
\theoremstyle{definition}
\newtheorem{assumption}{Assumption}

\newtheorem{proposition}{Proposition}
\newtheorem{remark}{Remark}

\newtheorem{theorem}{Theorem}

\title{Why Smooth Stability Assumptions Fail for ReLU Learning}

\author{ %\href{https://orcid.org/0000-0000-0000-0000}{\includegraphics[scale=0.06]{orcid.pdf}\hspace{1mm}
Ronald Katende %\thanks{Use footnote for providing further information about author (webpage, alternative address)---\emph{not} for acknowledging funding agencies.} 
\\
Department of Mathematics\\
Kabale University\\
Kikungiri Hill, Katuna Road, 317, Kabale, Uganda\\
\texttt{rkatende92@gmail.com} \\
}

\date{}

\renewcommand{\headeright}{}
\renewcommand{\undertitle}{}
\renewcommand{\shorttitle}{Why Smooth Stability Assumptions Fail for ReLU Learning}

\hypersetup{
pdftitle={Why Smooth Stability Assumptions Fail for ReLU Learning},
pdfauthor={Ronald Katende},
pdfkeywords={ReLU networks,  
	Nonsmooth optimization, 
	Stability analysis, 
	Generalized derivatives,
	Learning dynamics},
}

\begin{document}
\maketitle

\begin{abstract}
Stability analyses of modern learning systems are frequently derived under smoothness assumptions that are violated by ReLU-type nonlinearities. In this note, we isolate a minimal obstruction by showing that no uniform smoothness-based stability proxy such as gradient Lipschitzness or Hessian control can hold globally for ReLU networks, even in simple settings where training trajectories appear empirically stable. We give a concrete counterexample demonstrating the failure of classical stability bounds and identify a minimal generalized derivative condition under which stability statements can be meaningfully restored. The result clarifies why smooth approximations of ReLU can be misleading and motivates nonsmooth-aware stability frameworks.

\end{abstract}

% keywords can be removed
 % \and Second keyword \and More}

\section{Introduction}
\label{sec:intro}

Rectified linear units (ReLU) are the dominant activation function in contemporary deep learning architectures due to their computational simplicity and favorable optimization behavior \cite{GoodfellowBengioCourville2016}.
Despite their intrinsic nonsmoothness, analytic studies of stability and generalization for ReLU networks are frequently conducted using smoothness-based proxies, including global Lipschitz-gradient assumptions, Hessian-based bounds, or smooth surrogate activations \cite{HardtRechtSinger2016,NeyshaburBhojanapalliMcAllester2018}.
These assumptions enable tractable analysis but are not satisfied by ReLU networks at the level of the parameter-to-gradient map \cite{Clarke1990,RockafellarWets1998}.

In this note, a \emph{smoothness-based stability proxy} refers to the assumption that the parameter-to-gradient map $\theta \mapsto \nabla_\theta \mathcal{L}(\theta)$ is globally Lipschitz, so that small perturbations in parameters induce proportionally small changes in the update direction and hence in the resulting learning trajectory \cite{HardtRechtSinger2016,BousquetElisseeff2002}.

The purpose of this note is to isolate a precise obstruction.
We show that smoothness-based stability proxies cannot hold uniformly for ReLU learning dynamics, even for shallow architectures and smooth loss functions \cite{Clarke1990}.
The obstruction is structural rather than technical and arises from activation boundary crossings inherent to piecewise-linear nonlinearities \cite{RockafellarWets1998}.
Although such crossings occur on sets of measure zero in parameter space, they are dynamically unavoidable along generic gradient-based training trajectories \cite{LeeSimchowitzJordanRecht2016}.

The scope of this note is deliberately narrow.
Specifically, we establish the following:
\begin{enumerate}[label=(\roman*), leftmargin=*]
	\item a concrete counterexample demonstrating the failure of uniform smoothness-based stability bounds for ReLU networks \cite{Clarke1990,RockafellarWets1998}; and
	\item a minimal generalized derivative condition under which stability statements can be meaningfully formulated in nonsmooth learning systems \cite{Clarke1990}.
\end{enumerate}
We do not introduce new activation functions, training algorithms, or regularization schemes.
Rather, the goal is to clarify the limitations of commonly used analytic proxies and to delineate the boundary of their applicability \cite{BousquetElisseeff2002}.

\section{Setup and Notation}

We consider parameterized models of the form
\[
f(x;\theta) = W_2 \, \sigma(W_1 x),
\]
where $\sigma(z) = \max\{0,z\}$ is applied elementwise and $\theta = (W_1,W_2)$ denotes the collection of parameters \cite{GoodfellowBengioCourville2016}.
The loss functional $\mathcal{L}$ is assumed to be continuously differentiable with respect to the model output, as is standard in empirical risk minimization \cite{BousquetElisseeff2002}.

Training dynamics are given by gradient-based updates
\[
\theta_{t+1} = \theta_t - \eta \nabla_\theta \mathcal{L}(\theta_t),
\]
with step size $\eta>0$, as in classical analyses of algorithmic stability \cite{HardtRechtSinger2016}.
Throughout, stability refers to the sensitivity of the learning trajectory $\{\theta_t\}_{t\ge0}$ to perturbations in initialization or data \cite{BousquetElisseeff2002}.

We emphasize that, while $\nabla_\theta f(x;\theta)$ exists almost everywhere, it is discontinuous across activation boundaries \cite{Clarke1990}.
As a consequence, higher-order smoothness properties required by classical stability analyses, such as globally bounded Hessians or Lipschitz gradients, fail to hold for ReLU networks, even when the loss function itself is smooth \cite{RockafellarWets1998,NeyshaburBhojanapalliMcAllester2018}.

\section{Smoothness obstructions and a nonsmooth substitute}
\label{sec:smoothness-obstruction}

\subsection{Setting and smoothness proxy under test}
\label{subsec:setting}

Let $\sigma:\mathbb{R}\to\mathbb{R}$ be the ReLU, $\sigma(u)=\max\{0,u\}$, and consider the one-neuron model
\begin{equation}
	\label{eq:model-1d}
	f(x;\theta)=\sigma(wx+b),\qquad \theta:=(w,b)\in\mathbb{R}^2.
\end{equation}
Fix one datum $(x,y)\in\mathbb{R}\times\mathbb{R}$ and the smooth scalar loss
\begin{equation}
	\label{eq:loss}
	\mathcal{L}(\theta):=\frac12\big(f(x;\theta)-y\big)^2,
\end{equation}
so $\mathcal{L}$ is continuously differentiable in its output argument, but $\theta\mapsto f(x;\theta)$ is only piecewise-$C^\infty$ and changes regime across the activation boundary $\{wx+b=0\}$ \cite{Clarke1990}.

A common smoothness proxy used in stability and step-size analysis is global $L$-smoothness of $\mathcal{L}$ in parameters, i.e.,
\begin{equation}
	\label{eq:L-smooth}
	\|\nabla \mathcal{L}(\theta)-\nabla \mathcal{L}(\theta')\|\le L\|\theta-\theta'\|
	\quad\forall \theta,\theta'\in\mathbb{R}^2,
\end{equation}
equivalently: $\nabla \mathcal{L}$ is globally Lipschitz (and the Hessian exists a.e. and is uniformly bounded in operator norm when interpreted classically) \cite{Nesterov2004}.

Importantly, \eqref{eq:L-smooth} is a \emph{global} condition; local smoothness may hold within a fixed activation region, but uniform control fails once trajectories cross activation boundaries.

We show that \eqref{eq:L-smooth} cannot hold for ReLU even in \eqref{eq:model-1d}--\eqref{eq:loss}.

\subsection{Nonexistence of a global Lipschitz constant for the gradient map}
\label{subsec:failure}

\begin{theorem}[Failure of global $L$-smoothness for ReLU parameterizations]
	\label{thm:grad-not-lip}
	Fix any $x\neq 0$ and any $y\neq 0$. For $\mathcal{L}$ defined by \eqref{eq:model-1d}--\eqref{eq:loss}, there is no finite $L$ such that \eqref{eq:L-smooth} holds on $\mathbb{R}^2$. In particular, $\theta\mapsto\nabla \mathcal{L}(\theta)$ is not globally Lipschitz, hence any argument that requires a uniform bound on $\|\nabla^2\mathcal{L}(\theta)\|$ in the classical smooth sense cannot be valid globally \cite{Nesterov2004}.
\end{theorem}

\begin{proof}
	Let $s(\theta):=wx+b$. If $s(\theta)>0$, then $f(x;\theta)=s(\theta)$ and
	\[
	\mathcal{L}(\theta)=\frac12(s(\theta)-y)^2,
	\quad
	\nabla \mathcal{L}(\theta)=(s(\theta)-y)\nabla s(\theta)=(s(\theta)-y)(x,1).
	\]
	If $s(\theta)<0$, then $f(x;\theta)=0$ and $\mathcal{L}(\theta)=\frac12y^2$, hence
	\[
	\nabla \mathcal{L}(\theta)=(0,0)
	\qquad\text{for all }\theta\text{ with }s(\theta)<0,
	\]
	because $\mathcal{L}$ is locally constant on the open half-space $\{s<0\}$.
	
	Fix any $\theta_0$ on the boundary $s(\theta_0)=0$. Choose $\varepsilon>0$ and take
	\[
	\theta^-(\varepsilon):=\theta_0-\varepsilon(x,1),\qquad
	\theta^+(\varepsilon):=\theta_0+\varepsilon(x,1).
	\]
	Then
	\[
	s(\theta^-(\varepsilon))=-\varepsilon\|(x,1)\|^2<0,\qquad
	s(\theta^+(\varepsilon))=+\varepsilon\|(x,1)\|^2>0,
	\]
	so $\nabla\mathcal{L}(\theta^-(\varepsilon))=(0,0)$, while
	\[
	\nabla\mathcal{L}(\theta^+(\varepsilon))
	=\big(\varepsilon\|(x,1)\|^2-y\big)(x,1).
	\]
	Therefore
	\[
	\|\nabla\mathcal{L}(\theta^+(\varepsilon))-\nabla\mathcal{L}(\theta^-(\varepsilon))\|
	=
	\big|\,\varepsilon\|(x,1)\|^2-y\,\big|\,\|(x,1)\|.
	\]
	On the other hand,
	\[
	\|\theta^+(\varepsilon)-\theta^-(\varepsilon)\| = 2\varepsilon\|(x,1)\|.
	\]
	Hence the ratio satisfies
	\[
	\frac{\|\nabla\mathcal{L}(\theta^+(\varepsilon))-\nabla\mathcal{L}(\theta^-(\varepsilon))\|}
	{\|\theta^+(\varepsilon)-\theta^-(\varepsilon)\|}
	=
	\frac{\big|\,\varepsilon\|(x,1)\|^2-y\,\big|}{2\varepsilon}.
	\]
	Letting $\varepsilon\downarrow 0$ yields
	\[
	\frac{\big|\,\varepsilon\|(x,1)\|^2-y\,\big|}{2\varepsilon}\;\to\;\frac{|y|}{2\varepsilon}\;=\;+\infty,
	\]
	so no finite global Lipschitz constant $L$ can exist for $\nabla \mathcal{L}$. This proves the theorem.
\end{proof}
\paragraph{Takeaway (global vs.\ local).}
Theorem~\ref{thm:grad-not-lip} rules out \emph{global} $L$-smoothness for ReLU learning. However, \emph{local} smoothness can hold on any fixed activation region (no boundary crossings), so empirical stability along a trajectory is compatible with the failure of any uniform global proxy \cite{Clarke1990,RockafellarWets1998}.

\begin{remark}[Structural nature of the obstruction]
	\label{rem:generalization}
	\begin{center}
		\fbox{\parbox{0.95\linewidth}{
				\textbf{Depth- and width-robust obstruction.}
				Theorem~\ref{thm:grad-not-lip} is not a peculiarity of the one-neuron model \eqref{eq:model-1d}.
				It uses only the fact that ReLU networks are \emph{piecewise affine} in parameters with \emph{activation boundaries} separating regions.
				In any multi-neuron or multi-layer ReLU network, whenever a training trajectory crosses such a boundary, there exist parameter pairs separated by $O(\varepsilon)$ that lie on opposite sides of the boundary and produce an $O(1)$ change in the parameter-to-gradient map.
				Hence no architecture-independent global Lipschitz bound for $\theta\mapsto\nabla_\theta\mathcal{L}(\theta)$ (and no global bounded-Hessian proxy) can hold for ReLU parameterizations \cite{Clarke1990,RockafellarWets1998}.
		}}
	\end{center}
\end{remark}Figure~\ref{fig:fig1} below provides a geometric visualization of the obstruction established in Theorem~\ref{thm:grad-not-lip}.
\begin{figure}[!h]
	\centering\includegraphics[width=0.8\linewidth]{fig}
	\caption{Schematic illustration of the smoothness obstruction for ReLU learning. An $O(\varepsilon)$ parameter perturbation across the activation boundary $wx+b=0$ induces an $O(1)$ jump in the gradient, ruling out any global Lipschitz or bounded-Hessian proxy.}
	\label{fig:fig1}
\end{figure}

\subsection{Why smooth surrogates do not transfer uniformly to ReLU}
\label{subsec:surrogates}

A standard workaround replaces $\sigma$ by a smooth approximation $\sigma_\beta$ (e.g., softplus), proves an $L_\beta$-smoothness bound for the surrogate loss, and then informally interprets it as a statement about the ReLU limit \cite{Nesterov2004}. The obstruction in Theorem~\ref{thm:grad-not-lip} manifests as $L_\beta\to\infty$ as the approximation sharpens.

\begin{proposition}[Surrogate smoothness constants must diverge]
	\label{prop:surrogate-diverges}
	Let $\{\sigma_\beta\}_{\beta\ge 1}$ be $C^2$ activations such that $\sigma_\beta(u)\to\sigma(u)$ pointwise for all $u\in\mathbb{R}$ and $\sup_{u\in\mathbb{R}}|\sigma_\beta'(u)|\le 1$. Consider the surrogate model $f_\beta(x;\theta)=\sigma_\beta(wx+b)$ and the loss $\mathcal{L}_\beta(\theta)=\frac12(f_\beta(x;\theta)-y)^2$ with fixed $x\neq 0$, $y\neq 0$. If $\nabla \mathcal{L}_\beta$ is globally Lipschitz with constant $L_\beta$, then necessarily $L_\beta\to\infty$ along any sequence $\beta\to\infty$ for which $\sigma_\beta\to\sigma$ holds.
\end{proposition}

\begin{proof}
	For $s(\theta)=wx+b$, a direct differentiation gives
	\[
	\nabla \mathcal{L}_\beta(\theta)=(\sigma_\beta(s(\theta))-y)\sigma_\beta'(s(\theta))\,(x,1),
	\]
	and wherever $\sigma_\beta$ is twice differentiable,
	\[
	\nabla^2\mathcal{L}_\beta(\theta)
	=
	\Big(
	\big(\sigma_\beta'(s)\big)^2 + (\sigma_\beta(s)-y)\sigma_\beta''(s)
	\Big)\,(x,1)(x,1)^\top,
	\quad s=s(\theta).
	\]
	Thus, for any $\theta$,
	\begin{equation}
		\label{eq:hess-lower}
		\|\nabla^2\mathcal{L}_\beta(\theta)\|
		\;\ge\;
		\big|(\sigma_\beta(s(\theta))-y)\sigma_\beta''(s(\theta))\big|\;\|(x,1)\|^2,
	\end{equation}
	since $(x,1)(x,1)^\top$ has operator norm $\|(x,1)\|^2$.
	
	Because $\sigma_\beta\to\sigma$ pointwise and $\sigma$ has a kink at $0$, any $C^2$ family $\sigma_\beta$ approximating $\sigma$ must concentrate curvature near $0$ in the sense that $\sup_u|\sigma_\beta''(u)|\to\infty$ along $\beta\to\infty$; otherwise the derivatives would remain equicontinuous and could not converge to the discontinuous derivative of $\sigma$ at $0$ (a standard compactness/Arzel\`a--Ascoli contradiction; see nonsmooth approximation discussion in \cite{Clarke1990,RockafellarWets1998}). In particular, there exist $u_\beta$ with $|u_\beta|\to 0$ such that $|\sigma_\beta''(u_\beta)|\to\infty$.
	
	Pick $\theta_\beta$ with $s(\theta_\beta)=u_\beta$. Since $u_\beta\to 0$, we have $\sigma_\beta(u_\beta)\to\sigma(0)=0$, so $(\sigma_\beta(u_\beta)-y)\to -y$, hence $|\sigma_\beta(u_\beta)-y|\ge |y|/2$ for all large $\beta$. Plugging into \eqref{eq:hess-lower} yields
	\[
	\|\nabla^2\mathcal{L}_\beta(\theta_\beta)\|
	\;\ge\;
	\frac{|y|}{2}\,|\sigma_\beta''(u_\beta)|\,\|(x,1)\|^2
	\;\to\;\infty.
	\]
	If $\nabla \mathcal{L}_\beta$ were globally Lipschitz with constant $L_\beta$, then (by classical smooth calculus) $\|\nabla^2\mathcal{L}_\beta(\theta)\|\le L_\beta$ wherever the Hessian exists \cite{Nesterov2004}. Therefore $L_\beta\to\infty$.
\end{proof}

\subsection{A minimal nonsmooth regularity that matches the geometry}
\label{subsec:minimal}

Theorem~\ref{thm:grad-not-lip} rules out \eqref{eq:L-smooth} as a \emph{global} proxy. A compatible substitute is to work with generalized derivatives and formulate stability in terms of \emph{set-valued} Jacobians that remain bounded along the trajectory \cite{Clarke1990,RockafellarWets1998}.

Let $F:\mathbb{R}^d\to\mathbb{R}^d$ be locally Lipschitz. The Clarke generalized Jacobian $\partial_C F(\theta)$ is the convex hull of all limits of $\nabla F(\theta_n)$ along differentiability points $\theta_n\to\theta$ \cite{Clarke1990}. It is a nonempty compact convex set for locally Lipschitz $F$ \cite{Clarke1990}.

\begin{assumption}[Uniform bounded generalized Jacobian along the iterates]
	\label{ass:bounded-clarke}
	For the update map $T(\theta):=\theta-\eta \nabla \mathcal{L}(\theta)$, assume $T$ is locally Lipschitz on a region $\mathcal{D}\subset\mathbb{R}^d$ containing the iterates, and
	\begin{equation}
		\label{eq:bounded-clarke}
		\sup_{\theta\in\mathcal{D}}\;\sup_{A\in\partial_C T(\theta)}\|A\|\;\le\;\rho\;<\;\infty.
	\end{equation}
\end{assumption}
\noindent\emph{Interpretation.}
For ReLU networks, $T$ is locally Lipschitz on any region $\mathcal{D}$ that stays a positive distance away from activation boundaries; thus Assumption~\ref{ass:bounded-clarke} is inherently \emph{path/region-restricted} and is not expected to hold globally when boundary crossings occur \cite{Clarke1990,RockafellarWets1998}.

\begin{proposition}[A nonsmooth contraction criterion for perturbation stability]
	\label{prop:nonsmooth-contraction}
	Under Assumption~\ref{ass:bounded-clarke}, for any $\theta,\theta'\in\mathcal{D}$,
	\begin{equation}
		\label{eq:lipschitz-T}
		\|T(\theta)-T(\theta')\|\le \rho\,\|\theta-\theta'\|.
	\end{equation}
	In particular, if $\rho<1$ then the iteration $\theta_{t+1}=T(\theta_t)$ is a contraction on $\mathcal{D}$ and satisfies
	\[
	\|\theta_t-\theta_t'\|\le \rho^t\|\theta_0-\theta_0'\|
	\quad\text{whenever both trajectories remain in }\mathcal{D}.
	\]
\end{proposition}

\begin{proof}
	For locally Lipschitz maps, Clarke's mean value inequality gives
	\[
	\|T(\theta)-T(\theta')\|
	\le
	\sup_{z\in[\theta,\theta']}\;\sup_{A\in\partial_C T(z)}\|A\|\;\|\theta-\theta'\|,
	\]
	where $[\theta,\theta']$ denotes the line segment between $\theta$ and $\theta'$ \cite{Clarke1990}. Using \eqref{eq:bounded-clarke} yields \eqref{eq:lipschitz-T}. The contraction conclusion is then standard \cite{RockafellarWets1998}.
\end{proof}

\paragraph{Diagnostic message.}
Theorem~\ref{thm:grad-not-lip} says global smoothness fails because kink-crossings create $O(1)$ gradient jumps at $O(\varepsilon)$ parameter changes. Proposition~\ref{prop:nonsmooth-contraction} shows the correct object to control is not a classical Hessian bound, but the operator norms of generalized Jacobians of the \emph{update map} along the region actually visited by training \cite{Clarke1990,RockafellarWets1998}.

\section{Implications}

The results establish that smoothness-based stability proxies commonly invoked in analyses of ReLU learning systems cannot hold globally.
In particular, any argument that relies on uniform Lipschitz continuity of the gradient or uniformly bounded second derivatives is invalid once activation boundary crossings are admitted.
This invalidity is structural and persists independently of network depth, loss smoothness, or empirical training behavior.

At the same time, the obstruction identified here does not imply instability of ReLU-trained models.
Rather, it shows that classical smooth analytic surrogates fail to capture the correct geometric mechanism governing sensitivity in nonsmooth systems.
Consequently, empirical robustness of ReLU networks is not in contradiction with the absence of global smoothness bounds.

The analysis further implies that stability statements for ReLU architectures must be formulated using notions compatible with nonsmooth dynamics.
In particular, stability can only be meaningfully asserted through pathwise or region-restricted conditions that explicitly account for activation boundary geometry and generalized derivative behavior of the update map.
Smooth surrogates may provide local approximations but cannot serve as globally valid analytic proxies.

\section{Conclusion}

This note has identified a fundamental obstruction to smoothness-based stability analyses in ReLU learning systems.
By constructing an explicit counterexample and isolating the role of activation boundary crossings, we have shown that uniform smooth stability assumptions fail even in the simplest ReLU settings.

We have further demonstrated that this obstruction admits a precise nonsmooth formulation via generalized derivatives, yielding a minimal compatibility condition under which stability can be meaningfully discussed.
The results delineate a sharp boundary between analytically valid and invalid stability arguments and provide a rigorous baseline for the development of nonsmooth-aware stability frameworks for neural architectures.

\bibliographystyle{unsrt}
\bibliography{refs}
%\appendix

\end{document}